\documentclass[conference]{IEEEtran}
\IEEEoverridecommandlockouts
\usepackage{cite}
\usepackage{amsmath,amssymb,amsfonts}
\usepackage{algorithmic}
\usepackage{graphicx}
\usepackage{textcomp}
\usepackage{xcolor}

\usepackage{booktabs}
\usepackage{epsfig}
\usepackage{algorithm}
\usepackage{ntheorem}
\usepackage{multirow}

\usepackage{subfig}

\newtheorem{theorem}{\bf Theorem}

\def\BibTeX{{\rm B\kern-.05em{\sc i\kern-.025em b}\kern-.08em
    T\kern-.1667em\lower.7ex\hbox{E}\kern-.125emX}}
\begin{document}

\title{Conditional Distribution Learning for Graph Classification}

\author{\IEEEauthorblockN{1\textsuperscript{st} Jie~Chen}
\IEEEauthorblockA{\textit{College of Computer Science} \\
\textit{Sichuan University}\\
Chengdu, China \\
chenjie2010@scu.edu.cn}
\and
\IEEEauthorblockN{2\textsuperscript{rd} Hua~Mao}
\IEEEauthorblockA{\textit{Department of Computer and Information Sciences} \\
\textit{ Northumbria University}\\
Newcastle upon Tyne, U.K. \\
hua.mao@northumbria.ac.uk}
\and

\IEEEauthorblockN{3\textsuperscript{th} Chuanbin~Liu$^{\ast}$}
\IEEEauthorblockA{\textit{School of Economics and Management} \\
\textit{China University of Petroleum (Beijing)}\\
Beijing, China \\
liuchuanbingl@163.com}
\and
\IEEEauthorblockN{4\textsuperscript{nd} Zhu~Wang$^{\ast}$ \thanks{*Corresponding authors}}
\IEEEauthorblockA{\textit{Law School} \\
\textit{Sichuan University}\\
Chengdu, China \\
wangzhu@scu.edu.cn}
\and
\IEEEauthorblockN{5\textsuperscript{th} Xi Peng}
\IEEEauthorblockA{\textit{College of Computer Science} \\
\textit{Sichuan University}\\
Chengdu, China \\
pengx.gm@gmail.com}
}

\maketitle

\begin{abstract}
Graph contrastive learning (GCL) has shown promising performance in semisupervised graph classification. However, existing studies still encounter significant challenges in GCL. First, successive layers in graph neural network (GNN) tend to produce more similar node embeddings, while GCL aims to increase the dissimilarity between negative pairs of node embeddings. This inevitably results in a conflict between the message-passing mechanism of GNNs and the contrastive learning of negative pairs via intraviews. Second, leveraging the diversity and quantity of data provided by graph-structured data augmentations while preserving intrinsic semantic information is challenging. In this paper, we propose a self-supervised conditional distribution learning (SSCDL) method designed to learn graph representations from graph-structured data for semisupervised graph classification. Specifically, we present an end-to-end graph representation learning model to align the conditional distributions of weakly and strongly augmented features over the original features. This alignment effectively reduces the risk of disrupting intrinsic semantic information through graph-structured data augmentation. To avoid conflict between the message-passing mechanism and contrastive learning of negative pairs, positive pairs of node representations are retained for measuring the similarity between the original features and the corresponding weakly augmented features. Extensive experiments with several benchmark graph datasets demonstrate the effectiveness of the proposed SSCDL method.
\end{abstract}

\begin{IEEEkeywords}
graph representation learning, graph contrastive learning, conditional distribution learning, graph classification
\end{IEEEkeywords}

\section{Introduction}
\label{sec:Introduction}
Graph-structured data typically represent non-Euclidean data structures, which are often visualized as graphs with labels \cite{Kazemi2020RL}. The analysis of such data has attracted considerable attention in various applications, e.g., social network recommendation \cite{Fan2022GNNF}, molecule classification \cite{Wang2022MCL}, and traffic flow analysis \cite{Li2021STFG}. In graph classification tasks, graph-structured data consists of multiple graphs, each with its own independent topology. In practice, obtaining large-scale labelled annotations for graph-structured data is often prohibitively costly or even impossible. Semisupervised graph classification aims to predict the labels of graphs via a small number of labeled graphs and many unlabeled graphs.

Numerous graph neural networks (GNNs) have been proposed for the analysis of graph-structured data, e.g., the graph convolutional network (GCN) \cite{Kipf2017GCN}, graph attention network (GAT) \cite{Veli2018GAT} and their variants. Through the message-passing mechanism, GNNs aggregate semantic information from neighboring nodes in the graph topology to central nodes. In graph classification tasks, GNNs aims to transform individual graphs into corresponding low-dimensional vector representations at the graph level \cite{Yue2022SSGC}. Consequently, GNNs are capable of integrating graph-structured data into comprehensive graph representations, preserving the intrinsic structure of graph-structured data \cite{Vignac2020GNN}. Furthermore, data augmentation for graph-structured data has been extensively investigated to enhance the generalizability of GNN models.

Inspired by the previous success of visual data augmentation \cite{Wang2023CL, Chen2023DMC, Chen2020CL, Oord2018RL}, graph contrastive learning (GCL) leverages the advantages of graph data augmentation to maximize the similarity between positive pairs and the dissimilarity between negative pairs \cite{Wu2021SSL}. For example, You \textit{et al.} \cite{You2020GCL} proposed a GCL framework that employs four types of augmentations to learn invariant graph representations across different augmented views, including node dropping, edge perturbation, attribute masking and subgraph extraction. Zhu \textit{et al.} \cite{Zhu2021CGL} proposed a graph contrastive representation learning method with adaptive augmentation that incorporates various priors for topological and semantic information of the graph. These GCL-based approaches effectively enhance the generalizability of graph representation learning models.

GCL for graph classification has raised significant interests recently \cite{Ju2024SSGC, Han2022GDA, Yue2022SSGC}. For example, Han \textit{et al.} \cite{Han2022GDA} adopted $\mathcal{G}$-Mixup to augment graphs for graph classification. Yue \textit{et al.} \cite{Yue2022SSGC} proposed a graph label-invariant augmentation (GLIA) method that adds perturbations to the node embeddings of graphs for creating augmented graphs. However, these approaches have overlooked two limitations in GCL. First, different augmentations for graph-structured data often result in varying performances for GNNs \cite{Yue2022SSGC}. For example, heavy edge perturbation often disrupts the intrinsic semantic information contained in the graph-structured data. Therefore, a significant challenge is how to leverage the diversity and quantity provided by various graph-structured data augmentations while preserving the intrinsic semantic information. Second, successive GNN layers tend to make node representations more similar due to the message-passing mechanism \cite{Chen2020OSP}. In contrast, GCL aims to enhance the dissimilarity of negative pairs of node embeddings through mutual information maximization \cite{Yue2022SSGC}. This inevitably results in a potential conflict between message-passing mechanism of GNNs and the contrastive learning of negative pairs in intraviews.

In this paper, we propose a conditional distribution learning (CDL) method that learns graph representations from graph-structured data for semisupervised graph classification. We define slight perturbations to node attributes or graph topology as weak augmentations, whereas significant perturbations are referred to as strong augmentations. We first present an end-to-end graph representation learning model to align the conditional distributions of weakly and strongly augmented features over the original features. Aligning the conditional distributions effectively reduces the risk of disrupting intrinsic semantic information when strong augmentation is applied to the graph-structured data. Moreover, only positive pairs of node representations are retained for measuring the similarity between the original features and their corresponding weakly augmented features. From the perspective of the mutual information loss, we explain how the conflict between the message-passing mechanism of GNNs and the contrastive learning of negative pairs of node representations can be avoided. Additionally, we introduce a semisupervised learning scheme that includes a pretraining stage and a fine-tuning stage for semisupervised graph classification. Extensive experimental results with benchmark graph datasets demonstrate that the proposed CDL method is highly competitive with state-of-the-art graph classification methods.

Our major contributions are summarized as follows.
\begin{itemize}
\item We present an end-to-end graph representation learning model that takes advantage of both weak and strong augmentations of graph-structured data to learn graph representations for semisupervised graph classification.
\item Conditional distribution learning is introduced to characterize the consistent conditional distributions of weakly and strongly augmented node embeddings over the original node embeddings.
\item The similarity loss function is introduced to alleviate the potential conflict between the message-passing mechanism of the GNN and the contrastive learning of negative pairs in GCL.
\end{itemize}

\section{Preliminaries}
\label{sec:prelims}

\subsection{Problem Formulation}
Given a graph $\mathcal{G} = \left( {\mathcal{V}, \mathcal{E}} \right)$, $\mathcal{V}$ and $\mathcal{E}$ represent the sets of nodes and edges, respectively. Let $\mathbf{X} = \left[ {{\mathbf{x}_1},{\mathbf{x}_2},...,{\mathbf{x}_n}} \right] \in {\mathbb{R}^{d \times n}}$ denote the feature matrix of nodes, where $d$ is the feature dimension, and $n$ is the number of nodes. Let $\mathbf{A} \in {\mathbb{R}^{n \times n}}$ be the adjacency matrix of $\mathcal{G}$. If there exists an edge from node $i$ to node $j$, then $A_{ij}$ = 1; otherwise, $A_{ij}$ = 0. The degree matrix $\mathbf{D}$ is defined as $\mathbf{D} = diag\left[ {{d_1},{d_2},...,{d_n}} \right] \in {\mathbb{R}^{n \times n}}$, and its diagonal elements are ${d_i} = \sum\nolimits_{{v_j} \in V} {{A_{ij}}}$. Given an undirected graph $\mathcal{G}$, let $\widetilde {\mathbf{A}} = \mathbf{A} + \mathbf{I}$ denote the adjacency matrix of $\mathcal{G}$ with added self-loops. The diagonal degree matrix ${\widetilde {\mathbf{D}}}$ is defined as ${\widetilde D_{ii}} = \sum\nolimits_j {{{\widetilde A}_{ij}}}$. Given a set of graphs $\mathcal{G}_s = \left\{ {\left( {{\mathcal{G}_i},{y_i}} \right)} \right\}_{i = 1}^{{n_g}}$ with $\mathbf{X}_i \in {\mathbb{R}^{d \times n_i}}$ and $c$ categories, the purpose of graph classification is to learn a function $f$ that can map each graph to its corresponding label, where $n_g$ represents the number of graphs, $n_i$ denotes the number of nodes in ${\mathcal{G}_i}$,  and ${y_i}$ is the label of ${\mathcal{G}_i}$.

\subsection{Graph Classification Techniques}
GNNs have emerged as effective methods for graph classification \cite{Yue2022SSGC, Han2022GDA}. Without loss of generalizability, the hidden representation of node $j$ of a graph $\mathcal{G}_i$ is iteratively updated by aggregating information from its neighbors $\mathcal{N} \left( j \right)$ at the $l$-th layer, i.e.,
\begin{equation}\label{eq:agg}
\begin{split}
& \mathbf{z}_j^{\left( l \right)} = {f_{a}}\left( {\mathbf{h}_k^{\left( {l - 1} \right)}:k \in \mathcal{N}\left(j \right)} \right) \\
& \mathbf{h}_j^{\left( l \right)} = {f_{c}}\left( {\mathbf{h}_j^{\left( {l - 1} \right)},\mathbf{z}_j^{\left( l \right)}} \right)
\end{split}
\end{equation}
where ${f_{a}}{\left(  \cdot  \right)}$ is an aggregation function, $\mathcal{N}\left( i \right)$ represents the neighbors of node $i$, and ${f_{c}}{\left(  \cdot  \right)}$ combines the hidden representations of the neighbors and itself at the previous layer. For graph classification, GNNs ultimately apply a readout function to aggregate the hidden representations of nodes into a graph-level representation at the final layer, i.e.,
\begin{equation}\label{eq:readout}
\begin{split}
&{\mathbf{h}_{\mathcal{G}_i}} = {f_{readout}}\left( {\mathbf{h}_j^{\left( L \right)}:j \in \mathcal{V}\left( \mathcal{G}_i \right)} \right) \\
\end{split}
\end{equation}
where $L$ denotes the index of the final layer \cite{Feng2022MPGNN}. The graph-level representation ${\mathbf{h}_{\mathcal{G}_i}}$ can be employed to predict the label of $\mathcal{G}_i$, i.e.,
\begin{equation}\label{eq:sm}
\begin{split}
\mathbf{q}_i = {softmax} \left( {{\mathbf{h}_{\mathcal{G}_i}}} \right)
\end{split}
\end{equation}
where ${softmax}\left(  \cdot  \right)$ is applied to transform an input vector into a probability distribution for graph classification tasks.

\begin{figure*}
\centering
\includegraphics[width=0.7\linewidth]{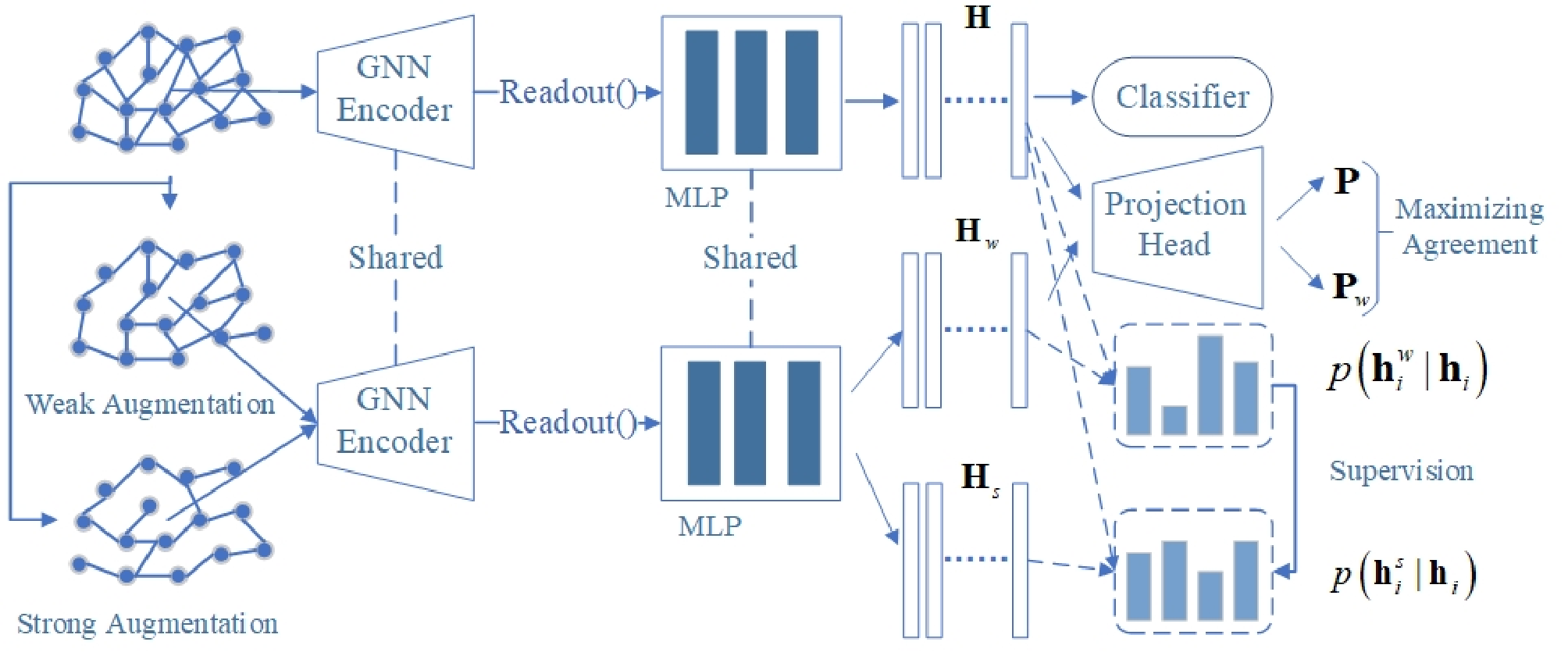}
\caption{Framework of the CDL model. The graph-level representations, $\mathbf{H}$, $\mathbf{H}_s$ and $\mathbf{H}_w$, are produced by a shared GNN encoder module using the graph-structured data, a strongly augmented view and a weakly augmented view, respectively. $p\left( {\mathbf{h}_i^s|{\mathbf{h}_i}} \right) $ and $p\left( {\mathbf{h}_i^w|{\mathbf{h}_i}} \right) $ represent the conditional distributions given the raw graph-level representation of the $i$th node. $\mathbf{P}$ and $\mathbf{P}_w$ denote the projected representations of the raw graph-structured data and a weakly augmented view, respectively.}
\label{fig:architecture} 
\end{figure*}

\section{Related Work}
\label{sec:work}
GCL has recently attracted significant interest. For example, the GCL framework exhibits a pretraining strategy of GNNs through maximizing the agreement between two augmented views of graphs \cite{You2020GCL}. Given $n$ graphs, $2n$ augmented graphs can be generated by applying a data augmentation technique, such as edge perturbation or attribute masking. The normalized temperature-scaled cross entropy loss (NT-Xent) \cite{Oord2018RL, Sohn2016DML} is adopted to maximize the consistency between positive pairs ${\mathbf{u}_i},{\mathbf{v}_i}$ compared with negative pairs, i.e.,
\begin{equation}\label{eq:clloss}
\begin{split}
{\mathcal{L}_i} =  - \log \frac{{f\left( {{\mathbf{u}_i},{\mathbf{v}_i}} \right)}}{{f\left( {{\mathbf{u}_i},{\mathbf{v}_i}} \right) + \sum\limits_{k \ne i}^n {f\left( {{\mathbf{u}_i},{\mathbf{v}_k}} \right) + \sum\limits_{k \ne i}^n {f\left( {{\mathbf{u}_i},{\mathbf{u}_k}} \right)} } }}
\end{split}
\end{equation}
where $f\left(  \cdot  \right) = \exp \left( {{{sim\left(  \cdot  \right)} \mathord{\left/
 {\vphantom {{sim\left(  \cdot  \right)} \tau }} \right.
 \kern-\nulldelimiterspace} \tau }} \right)$, $sim\left( {{\mathbf{u}_i},{\mathbf{v}_j}} \right) = \frac{{{\mathbf{u}_i}{\mathbf{v}_j}}}{{\left\| {{\mathbf{u}_i}} \right\|\left\| {{\mathbf{v}_j}} \right\|}}$ is the cosine similarity and $\tau$ is a temperature parameter. In GCL, a conflict arises in the message-passing mechanism of GNNs when combined with contrastive learning of negative pairs via intraviews. This is because certain samples simultaneously contribute to the gradients of both positive and negative pairs during message passing \cite{Ji2024RMP}.

Recent studies have been proposed to investigate semisupervised graph classification \cite{Ju2024SSGC, Yue2022SSGC, Zeng2021CSSL, Chen2019GNN}. These methods typically leverage a small number of labeled graphs and massive numbers of available unlabeled graphs to explore the semantic information of graph-structured data. Most existing GCL-based methods for graph classification tasks inherently employ contrastive learning of both positive and negative pairs \cite{Wang2024GCMAE, Yue2022SSGC}. This inevitably leads to a conflict between the message-passing mechanism of GNNs and the contrastive learning of negative pairs via intraviews. Additionally, data augmentation often disrupt the intrinsic semantic information contained in the graph-structured data \cite{Yue2022SSGC}, which makes enhancing the generalizability and robustness of the graph classification models challenging.

\section{Conditional Distribution Learning on Multiple Graphs}
\label{sec:sucdl}

\subsection{Framework overview }
Fig. \ref{fig:architecture} shows the framework of the CDL model, which consists of three modules: a shared GNN encoder module, a projection head module and a conditional distribution construction module.  The shared GNN encoder module learns graph-level representations, $\mathbf{H}$, $\mathbf{H}_s$ and $\mathbf{H}_w$, from the raw graph-structured data, a strongly augmented view and a weakly augmented view, respectively. These graph-level representations are obtained from the node embeddings of the graphs through a graph pooling layer. In particular, $\mathbf{H}$ is used for semisupervised graph classification tasks. The projection head module is designed to produce the projected representations of the raw graph-structured data augmentation and a weakly augmented view for GCL. The conditional distribution construction module constructs the conditional distributions of strongly and weakly augmented node embeddings given the original node embeddings.

\subsection{Conditional Distribution Learning}
We formally introduce the details of the proposed CDL method, including a semisupervised learning scheme for graph classification and the CDL model. In particular, the semisupervised training scheme for graph classification consists of two stages: a pretraining stage and a fine-tuning stage.

\subsubsection{Pretraining Stage}
When many unlabeled graphs are available, an intuitively designed pretraining framework is an effective approach for semisupervised graph classification. The pretraining framework of the CDL model includes a shared GNN encoder module of graph-structured data and a projection head module. The shared GNN encoder module consists of three components, including a GNN encoder component, a pooling layer component and a multilayer perceptron (MLP) component. In the shared GNN encoder module, we perform graph representation learning according to the graph classification techniques.

For simplicity, a GCN \cite{You2020GCL, Kipf2017GCN} is used as an example to describe the backbone of the GNN encoder module. The formula for the $l$th GCN layer in the GNN encoder is defined as follows:
\begin{equation}\label{eq:gcn}
\begin{split}
{\mathbf{H}^{(l)}} & = \sigma \left( {{{\widetilde {\mathbf{D}}}^{ - {1 \mathord{\left/
 {\vphantom {1 2}} \right.
 \kern-\nulldelimiterspace} 2}}}{\widetilde {\mathbf{A}}}{{\widetilde {\mathbf{D}}}^{ - {1 \mathord{\left/
 {\vphantom {1 2}} \right.
 \kern-\nulldelimiterspace} 2}}}{\mathbf{H}^{(l - 1)}}{\mathbf{W}^{(l)}}} \right) \\
\end{split}
\end{equation}
where ${\mathbf{W}^{(l)}}$ is a weight matrix of the $l$th layer, ${\mathbf{H}^{(l)}}$ represents a node-level embedding matrix with $\mathbf{H}^{(0)}=\mathbf{X}$, and $\sigma$ is a nonlinear activation function, e.g., ${\rm ReLU}\left(  \cdot  \right) = \max \left( {0, \cdot } \right)$. Then, we employ a global sum pooling layer as a READOUT function to obtain graph-level representations from the node-level embeddings, i.e.,
${\mathbf{H}_\mathcal{G}} = pooling\left( {\mathbf{H}^{(L)}} \right)$. Finally, the MLP component is composed of several fully connected layers, with computations defined as follows:
\begin{equation}\label{eq:fmlp}
\begin{split}
&\mathbf{H}= \mathbf{MLP}\left( {\mathbf{H}_\mathcal{G},{\mathbf{W}^{(l)}}} \right) \\
\end{split}
\end{equation}
where a nonlinear activation function, ${\rm ReLU}\left(  \cdot  \right)$, is applied for each linear layer. To improve the generalizability of the CDL model, we apply a batch normalization function in individual layers \cite{Ioffe2015BN}.

During the pretraining stage, the graph-level representations, $\mathbf{H}$ and $\mathbf{H}_w$, are produced by the shared GNN encoder module via the raw graph-structured data and a weakly augmented view, respectively. We further employ the projection head module \cite{Chen2020CL}, which consists of a two-layer MLP, to obtain the projections of $\mathbf{H}$ and $\mathbf{H}_w$ for contrastive learning, i.e.,
\begin{equation}\label{eq:proj}
\begin{split}
& \mathbf{P} = {\mathbf{W}^{\left( 2 \right)}}\sigma\left( {{\mathbf{W}^{\left( 1 \right)}}\mathbf{H}} \right), \\
& \mathbf{P}_w = {\mathbf{W}^{\left( 2 \right)}}\sigma\left( {{\mathbf{W}^{\left( 1 \right)}}\mathbf{H}_w} \right).
\end{split}
\end{equation}
The pretraining framework of the CDL model aims to ensure the consistency of the intrinsic semantic information between a weakly augmented view and the original graph-structured data. Therefore, we further investigate the loss function employed during the pretraining stage.

\subsubsection{Conditional Distribution Learning}
When dealing with relatively small amounts of graph-structured data, GNN-based models often learn redundant features due to the message-passing and aggregation mechanisms. Graph-structured data augmentation aims to increase the diversity and quantity of graph-structured data, which can improve the generalizability and robustness of GCL-based models. To effectively  learn essential features from graph-structured data, GCL-based models apply strong augmentation to graph-structured data. However, strong augmentations, such as edge perturbation or attribute masking, often significantly perturb the original graph-structured data, potentially altering the intrinsic semantic information contained in the graph-structured data and thus resulting in limited generalizability of GCL-based models.

Inspired by recent advances of contrastive learning techniques in computer vision \cite{Wang2023CL},  we utilize a conditional distribution to enforce the consistency of intrinsic semantic information between an augmented view and the original graph-structured data. To minimize the risk of altering intrinsic semantic information with strong augmentation, we introduce a conditional distribution learning strategy to guide graph representation learning. This learning strategy explores the supervision information embedded in the original graph-structured data for weak and strong augmentations. We consider the distribution of relative similarity between a weakly augmented node embedding and the original node embedding. Specifically, the conditional distribution of $\mathbf{h}_i^w$ given $\mathbf{h}_i$ is defined as follows:
\begin{equation}\label{eq:ff}
\begin{split}
&p\left( {{\mathbf{h}}_i^w|{{\mathbf{h}}_i}} \right) = \frac{{p\left( {{\mathbf{h}}_i^w,{{\mathbf{h}}_i}} \right)}}{{p\left( {{{\mathbf{h}}_i}} \right)}} \\
 & = \frac{{\exp \left( {{{sim\left( {{\mathbf{h}}_i^w,{{\mathbf{h}}_i}} \right)} \mathord{\left/
 {\vphantom {{sim\left( {{\mathbf{h}}_i^w,{{\mathbf{h}}_i}} \right)} \tau }} \right.
 \kern-\nulldelimiterspace} \tau }} \right)}}{{\exp \left( {{{sim\left( {{\mathbf{h}}_i^w,{{\mathbf{h}}_i}} \right)} \mathord{\left/
 {\vphantom {{sim\left( {{\mathbf{h}}_i^w,{{\mathbf{h}}_i}} \right)} \tau }} \right.
 \kern-\nulldelimiterspace} \tau }} \right) + \sum\limits_{k = 1}^K {\exp \left( {{{sim\left( {{\mathbf{h}}_k^w,{{\mathbf{h}}_i}} \right)} \mathord{\left/
 {\vphantom {{sim\left( {{\mathbf{h}}_k^w,{{\mathbf{h}}_i}} \right)} \tau }} \right.
 \kern-\nulldelimiterspace} \tau }} \right)} }}
 \end{split}
\end{equation}
where $K$ denotes the number of negative samples chosen from a weakly augmented view, and $\mathbf{h}_i$ and $\mathbf{h}_i^w$ represent the corresponding original node embedding and weak node embedding, respectively. Additionally, the other conditional distributions of $\mathbf{h}_i^s$ given $\mathbf{h}_i$, i.e., $p\left( {\mathbf{h}_i^s|{\mathbf{h}_i}} \right)$, can be constructed similarly to $p\left( {\mathbf{h}_i^w|{\mathbf{h}_i}} \right)$. Thus, we construct the conditional distributions of weakly and strongly augmented node embeddings given the original node embeddings.

To maintain the consistency of intrinsic semantic information contained in the graph-structured data between weak and strong augmentations, we align the conditional distributions of weakly and strongly augmented features over the original features, represented as $p\left( {\mathbf{h}_i^w|{\mathbf{h}_i}} \right)$ and $p\left( {\mathbf{h}_i^s|{\mathbf{h}_i}} \right)$. Specifically, we present a metric named the distribution divergence to quantify the divergence between these two conditional distributions as follows:
\begin{equation}\label{eq:lossdd}
\begin{split}
{\mathcal{L}_d} =  - \frac{1}{n_g}\sum\limits_{i = 1}^{n_g} {p\left( {\mathbf{h}_i^w|{\mathbf{h}_i}} \right)\log \left( p\left( {\mathbf{h}_i^s|{\mathbf{h}_i}} \right) \right)}.
\end{split}
\end{equation}
By minimizing ${\mathcal{L}_d}$, the conditional distributions of weakly augmented node embeddings given the original node embeddings are employed to supervise the corresponding conditional distributions of strongly augmented node embeddings given the original node embeddings. The original node embeddings provide useful supervision information to ensure the consistency of the intrinsic semantic information between weak and strong augmentations. Consequently, this alignment enables the CDL model preserve intrinsic semantic information when both weak and strong augmentations are applied to graph-structured data.

The intrinsic semantic information within graph-structured data remains susceptible to distortion, even though weak augmentation is applied. We focus on how $ p\left( {\mathbf{h}_i^w|{\mathbf{h}_i}} \right)$ provides valuable supervision information for $p\left( {\mathbf{h}_i^s|{\mathbf{h}_i}} \right)$. GCL is considered an effective strategy for mitigating the potential risk of distortion during weak augmentation. Unfortunately, there is a conflict between the message-passing mechanism of GNNs and the contrastive learning of negative pairs in augmented views, as previously discussed. For example, a node embedding ${\mathbf{h}_i^w}$ is aggregated from the neighboring nodes $\left\{ {{\mathbf{h}_j^w}:j \in {\cal N}\left( i \right)} \right\}$ in the weakly augmented view. However, these neighboring nodes are considered negative for ${\mathbf{h}_i^w}$ when the loss function in Eq. \eqref{eq:clloss} is employed. To evaluate the similarity between a positive pair $\left( {\mathbf{h}_i^w, {\mathbf{h}_i}} \right)$, the similarity loss is formulated as follows:
\begin{equation}\label{eq:losssim}
\begin{split}
{\mathcal{L}_s} =  - \frac{1}{{{n_g}}}\sum\limits_{i = 1}^{{n_g}} {\log \frac{{\exp \left( {{{sim\left( {\mathbf{p}_i^w,{\mathbf{p}_i}} \right)} \mathord{\left/
 {\vphantom {{sim\left( {\mathbf{p}_i^w,{\mathbf{p}_i}} \right)} \tau }} \right.
 \kern-\nulldelimiterspace} \tau }} \right)}}{{\sum\limits_{j \ne i}^{{n_g}} {\exp \left( {{{sim\left( {\mathbf{p}_i^w,{\mathbf{p}_j}} \right)} \mathord{\left/
 {\vphantom {{sim\left( {\mathbf{p}_i^w,{\mathbf{p}_j}} \right)} \tau }} \right.
 \kern-\nulldelimiterspace} \tau }} \right)} }}}
\end{split}
\end{equation}
where $\mathbf{p}_i^w$, ${\mathbf{p}_i}$ and ${\mathbf{p}_j}$ are computed via Eq. \eqref{eq:proj}. Considering a pair of negative node embeddings $\left( {\mathbf{h}_i^w,{\mathbf{h}_j}} \right)$ $\left(j \ne i\right)$ in Eq. \eqref{eq:losssim}, $\mathbf{h}_i^w$ does not participate in the message-passing of the CDL model involving with ${\mathbf{h}_j}$, as ${\mathbf{h}_j}$ originates from the original view. This similarity loss helps mitigate the aforementioned conflict in graph representation learning, making it beneficial for preventing the CDL model from overfitting to graph representations. The loss function in Eq. \eqref{eq:losssim} is employed during the pretraining stage.

Applying strong augmentations to graph-structured data may distort its intrinsic semantic information. Similarly, weak augmentations can also introduce subtle distortions. Eq. \eqref{eq:lossdd} ensures consistency between weak and strong augmentations. Furthermore, Eq. \eqref{eq:losssim} aims to preserve the complete intrinsic semantic information of graph-structured data as much as possible under weak augmentations. This preservation makes the alignment of conditional distributions in Eq. \eqref{eq:lossdd} both meaningful and effective.

\begin{algorithm}[tb]
\caption{Optimization Procedure for CDL}
\label{alg:sscdl}
\textbf{Input}: The training graphs $\mathcal{G}_r = \left\{ {\left( {\mathcal{G}_i} \right)} \right\}_{i = 1}^{{n_g}}$ and the testing graphs $\mathcal{G}_t = \left\{ {\left( {\mathcal{G}_i} \right)} \right\}_{i = 1}^{{n_t}}$. \\
\textbf{Parameter}: The number of iterations $epochs$, parameters $\alpha > 0 $ and $\beta > 0$.\\
\textbf{Output}: The predicted labels of the testing graphs $\mathcal{G}_t$.
\begin{algorithmic}[1]
\STATE Construct a weakly augmented view $\mathcal{G}_w$ and a strongly augmented view $\mathcal{G}_s$ from $\mathcal{G}_r$;
\FOR{ ${t = 1}$ to $epochs$ }
\STATE Compute ${\mathbf{P}}$ and ${\mathbf{P}_w}$ via Eq. \eqref{eq:proj} using the unlabeled parts of $\mathcal{G}_r$ and $\mathcal{G}_w$, respectively;
\STATE Update the network by minimizing ${\mathcal{L}_{s}}$ in Eq. \eqref{eq:losssim}
\ENDFOR
\FOR{ ${t = 1}$ to $epochs$ }
\STATE Compute $\mathbf{q}_i$ via Eq. \eqref{eq:sm} for each labeled graph in $\mathcal{G}_r$;
\STATE Compute ${\mathbf{P}}$ and ${\mathbf{P}_w}$ via Eq. \eqref{eq:proj} using the labeled parts of $\mathcal{G}_r$ and $\mathcal{G}_w$, respectively;
\STATE Compute $p\left( {\mathbf{h}_i^w|{\mathbf{h}_i}} \right)$ and $p\left( {\mathbf{h}_i^s|{\mathbf{h}_i}} \right)$ via Eq. \eqref{eq:ff} using the labeled parts of $\mathcal{G}_r$, $\mathcal{G}_s$ and $\mathcal{G}_w$;
\STATE Update the network by minimizing ${\mathcal{L}}$ in Eq. \eqref{eq:ovloss};
\ENDFOR
\STATE Compute $\mathbf{q}_i$ via Eq. \eqref{eq:sm} for each testing graph in $\mathcal{G}_t$. \\
\end{algorithmic}
\end{algorithm}

\subsubsection{Fine-Tuning Stage}
During the fine-tuning stage, a small number of labeled graphs are employed to train the CDL model. We employ the cross-entropy loss function for graph classification, i.e.,
\begin{equation}\label{eq:ce}
\begin{split}
{\mathcal{L}_c} =  - \frac{1}{{{n_g}}}\sum\limits_{i = 1}^{{n_g}} {\sum\limits_{j = 1}^c {{y_{ij}}\log {q_{ij}}} }
\end{split}
\end{equation}
where ${\mathbf{y}_i}\in {\mathbb{R}^{c}}$ denotes the one hot encoding for the label of the $i$th graph, and ${q}_{ij}$ can be computed via Eq. \eqref{eq:sm}. We can obtain the graph-level representation $\mathbf{H}_s$ from a strongly augmented view via the shared GNN encoder module, similar to $\mathbf{H}$ and $\mathbf{H}_w$. The loss function of the CDL model in Eq. \eqref{eq:lossdd} is introduced in the fine-tuning stage. Instead of maximizing the mutual information between the strongly augmented view and the original graph-structured data, the weakly augmented view can provide useful information to bridge the gap by minimizing the distribution divergence ${\mathcal{L}_d}$. Therefore, the overall loss of the proposed CDL model consists of three main components:
\begin{equation}\label{eq:ovloss}
\begin{split}
{\mathcal{L}} = {\mathcal{L}_c} + \alpha {\mathcal{L}_s} + \beta {\mathcal{L}_d}
\end{split}
\end{equation}
where $\alpha$ and $\beta$ are tradeoff hyperparameters. Algorithm \ref{alg:sscdl} summarizes the entire optimization procedure of the proposed CDL method.

\subsection{Theoretical Justification}

\subsubsection{Why is Pretraining Needed in CDL?}
Let two random variables $\mathbf{U} \in {\mathbb{R}^{d'}}$ and $\mathbf{V}\in {\mathbb{R}^{d'}}$ be node embeddings corresponding to a weakly augmented view and the original view, respectively, where ${d'}$ is the dimensionality of the node embeddings. The loss function of the CDL model in Eq. \eqref{eq:losssim} is a lower bound of mutual information (MI), i.e.,
\begin{equation}\label{eq:mi1}
\begin{split}
I\left( {\mathbf{U}; \mathbf{V}} \right)  \ge \log \left( {{n_g} - 1} \right) - {\mathcal{L}_s}.
\end{split}
\end{equation}
Therefore, minimizing the loss function of the CDL model in Eq. \eqref{eq:losssim} equivalent to maximizing a lower bound on $I\left( {\mathbf{U}; \mathbf{V}} \right) $. The proof is similar to that of minimizing the InfoNCE loss \cite{Oord2018RL}. By pretraining, the intrinsic semantic information is preserved as much as possible when applying weak augmentation to the graph-structured data. This explains why $ p\left( {\mathbf{h}_i^w|{\mathbf{h}_i}} \right)$ can provide valuable supervision information for $p\left( {\mathbf{h}_i^s|{\mathbf{h}_i}} \right)$ in the CDL model.

\begin{table*}[t]
\caption{The graph classification results (average accuracy (\%) $\pm$ standard deviation (\%)) with eight benchmark graph datasets.}
\vskip -0.1in
\label{tb:results}
\setlength{\tabcolsep}{3pt}
\begin{center}
\begin{small}
\begin{tabular}{l|c|ccccccccr|}
\toprule
 Label  & Methods & MUTAG & PROTEINS & IMDB-B & NCI1 & RDT-B & RDT-M5K & COLLAB & GITHUB \\
\midrule
\multirow{8}{*}{30\%} & GCN & 81.52$\pm$11.12 & 67.84$\pm$3.46 & 71.90$\pm$5.59 & 71.48$\pm$1.65 & 86.65$\pm$1.86 & 51.81$\pm$2.81 & 71.04$\pm$2.98 & 65.65$\pm$1.07 \\
~ & GAT & 82.02$\pm$9.58 & 63.25$\pm$3.31 & 72.60$\pm$4.40 & 64.94$\pm$2.83 & 79.30$\pm$4.18 & 49.83$\pm$1.85 & 71.42$\pm$5.82 & 65.28$\pm$0.83 \\
~ & GCL & 84.65$\pm$8.32 & 74.95$\pm$5.54 & 72.00$\pm$3.59 & 76.08$\pm$2.12 & \underline{90.40$\pm$1.74} & 55.01$\pm$1.39 & 77.56$\pm$2.49 & 64.70$\pm$1.23  \\
~ & GLIA &  \underline{87.25$\pm$8.44} &   \underline{75.12$\pm$4.54} &  \underline{73.30$\pm$4.45} &  \underline{76.72$\pm$2.20} &  90.35$\pm$1.63 &  \underline{55.09$\pm$1.45} &  \underline{77.94$\pm$1.81} &  \underline{68.61$\pm$1.24} \\
~ & G-Mixup & 82.49$\pm$8.49 &  72.06$\pm$3.63 & 72.40$\pm$5.58 & 75.88$\pm$2.35 & 89.50$\pm$2.90 & 51.45$\pm$2.78 & 73.92$\pm$1.37 & 65.56$\pm$1.59  \\
~ & GCMAE &  70.23$\pm$7.46 & 69.46$\pm$3.41 & 72.70$\pm$4.47 & 61.97$\pm$1.73 & 76.60$\pm$2.03 & - & 72.14$\pm$1.66 & - \\
~ & GRDL & 83.45$\pm$9.12 & 69.55$\pm$3.80 & 72.50$\pm$4.32 & 68.49$\pm$0.12 & - & - & 71.64$\pm$6.77 & - \\
~ & CDL  & \textbf{89.36$\pm$6.14}  & \textbf{76.74$\pm$4.51} & \textbf{74.60 $\pm$ 4.25} & \textbf{79.37 $\pm$1.62} & \textbf{91.15$\pm$1.76} & \textbf{55.31$\pm$1.65} & \textbf{79.44$\pm$1.82} & \textbf{70.15$\pm$1.27} \\
\midrule
\multirow{8}{*}{50\%} & GCN & 83.10$\pm$8.83 & 69.27$\pm$3.17 & 72.40$\pm$4.99 & 72.80$\pm$1.93 & 87.30$\pm$3.54 & 52.59$\pm$3.19 & 73.98$\pm$1.58 & 66.55$\pm$1.17 \\
~ & GAT &  84.06$\pm$8.64 & 67.12$\pm$5.09 & 73.10$\pm$3.18 & 66.62$\pm$2.29 & 80.00$\pm$4.06 & 51.23$\pm$2.68 & 73.88$\pm$3.35 & 65.82$\pm$1.34 \\
~ & GCL & 85.09$\pm$8.19 & \underline{75.30$\pm$3.50} & 72.80$\pm$4.47 & 76.72$\pm$1.61 & 90.80$\pm$1.7 & 55.97$\pm$2.23 & 79.36$\pm$2.04 & 64.46$\pm$1.54 \\
~ & GLIA &  \underline{89.39$\pm$10.21} &  75.03$\pm$3.78 &  \underline{74.30$\pm$4.22} & 76.28$\pm$2.11 &  \underline{90.90$\pm$2.07} &  \underline{56.11$\pm$1.60} & \underline{79.66$\pm$1.22}  &  \underline{69.75$\pm$1.23} \\
~ & G-Mixup & 81.43$\pm$8.32 & 72.15$\pm$3.45 & 72.90$\pm$5.58 & \underline{77.02$\pm$2.20} & 89.80$\pm$1.70 & 52.27$\pm$2.09 & 74.30$\pm$1.83 & 66.14$\pm$1.45 \\
~ & GCMAE &  74.97$\pm$7.21 & 71.71$\pm$3.06 & 72.90$\pm$3.56 &  63.63$\pm$1.57 & 77.15$\pm$3.02 & - & 72.20$\pm$1.82 & - \\
~ & GRDL & 84.59$\pm$9.23 & 70.36$\pm$2.48 & 73.60$\pm$4.72 & 72.51$\pm$1.50 &  - & - & 73.04$\pm$11.74 & - \\
~ & CDL  &  \textbf{89.94$\pm$8.76} & \textbf{76.10$\pm$2.80} & \textbf{74.90$\pm$3.70} & \textbf{79.08$\pm$1.86} & \textbf{92.05$\pm$2.14} & \textbf{56.51$\pm$2.32} & \textbf{80.96$\pm$1.29} & \textbf{70.83$\pm$1.13} \\
\midrule
\multirow{8}{*}{70\%} & GCN & 84.21$\pm$11.37 & 69.28$\pm$4.72 & 72.80$\pm$4.61 & 75.09$\pm$2.29 & 88.50$\pm$2.59 & 53.59$\pm$2.19 & 74.84$\pm$1.72 & 67.50$\pm$1.75  \\
~ & GAT &  84.68$\pm$8.32 & 68.46$\pm$4.89 & 73.30$\pm$4.00 & 67.62$\pm$2.74 & 81.05$\pm$3.83 & 51.53$\pm$2.71 & 72.18$\pm$1.43 & 66.04$\pm$1.38  \\
~ & GCL & 88.80$\pm$5.87 &  \underline{76.20$\pm$4.32}  & 73.70$\pm$5.60 & 78.47$\pm$2.60 & 91.25$\pm$1.72 &  56.81$\pm$1.55 & 80.82$\pm$1.53 & 65.54$\pm$1.03 \\
~ & GLIA &  \underline{89.42$\pm$7.45} & 75.75$\pm$4.82 &  \underline{74.50$\pm$3.08} &  79.29$\pm$2.13 &  \underline{91.40$\pm$1.96} &  \underline{56.09$\pm$1.56} &  \underline{81.18$\pm$1.67} &  \underline{69.95$\pm$0.77} \\
~ & G-Mixup & 84.06$\pm$6.63 & 72.07$\pm$5.25 & 73.10$\pm$3.60 & \underline{79.71$\pm$2.48} & 89.70$\pm$2.63 & 52.31$\pm$2.22 & 75.72$\pm$2.78 & 67.58$\pm$1.76 \\
~ & GCMAE & 76.05$\pm$5.92 & 72.37$\pm$2.45 & 73.60$\pm$5.10 & 64.89$\pm$1.99 & 78.35$\pm$3.03 & - & 72.18$\pm$1.43 & - \\
~ & GRDL & 85.67$\pm$7.02 & 71.53$\pm$2.85 & 73.30$\pm$5.83 & 72.51$\pm$1.50 &  - & - & 74.06$\pm$10.33 & - \\
~ & CDL & \textbf{89.91$\pm$7.30} & \textbf{77.27$\pm$3.62} & \textbf{74.90$\pm$5.63} & \textbf{82.36$\pm$1.52} & \textbf{92.35$\pm$1.63} & \textbf{56.65$\pm$1.87} & \textbf{82.36$\pm$1.72} & \textbf{71.06$\pm$1.17} \\
\bottomrule
\end{tabular}
\end{small}
\end{center}
\vskip -0.2in
\end{table*}

\begin{table*}[t]
\caption{Ablation study of the main components in Eq. \eqref{eq:ovloss} with eight benchmark graph datasets.}
\vskip -0.1in
\label{tb:ablation}
\setlength{\tabcolsep}{2pt}
\begin{center}
\begin{small}
\begin{tabular}{l|c|cccccccccccr|}
\toprule
 Label & Methods & ${\mathcal{L}_s}$ & ${\mathcal{L}_c}$ & ${\mathcal{L}_d}$ & MUTAG & PROTEINS & IMDB-B & NCI1 & RDT-B & RDT-M5K & COLLAB & GITHUB \\
\midrule
\multirow{3}{*}{30\%} & CDL$_{\textbf{cl}}$  & & \checkmark & &  87.22$\pm$8.21 & 74.49$\pm$4.24  & 70.70$\pm$4.22  & 78.25$\pm$1.63 & 90.15$\pm$1.45 & 54.55$\pm$1.76 & 78.64$\pm$1.86 & 68.44$\pm$0.56 \\
~& CDL$_{\textbf{ft}}$ &  \checkmark & \checkmark &  & 87.81$\pm$8.66 & 74.75$\pm$3.86 & 71.20$\pm$4.64  & 78.64$\pm$1.62 & 90.20$\pm$1.36 & 54.25$\pm$1.75 & 78.70$\pm$0.94 & 68.04$\pm$1.26 \\
~ & CDL & \checkmark& \checkmark& \checkmark & \textbf{89.36$\pm$6.14}  & \textbf{76.74$\pm$4.51} & \textbf{74.60 $\pm$ 4.25} & \textbf{79.37 $\pm$1.62} & \textbf{91.15$\pm$1.76} & \textbf{55.31$\pm$1.65} & \textbf{79.44$\pm$1.82} & \textbf{70.15$\pm$1.27} \\
\midrule
\multirow{3}{*}{50\%} & CDL$_{\textbf{cl}}$  & & \checkmark &  &  89.36$\pm$7.89  & 74.58$\pm$3.45 & 71.3$\pm$5.56  & 77.91$\pm$2.65 & 90.25$\pm$1.27 & 55.19$\pm$2.02 & 79.98$\pm$1.16 & 69.04$\pm$1.06 \\
~& CDL$_{\textbf{ft}}$ & \checkmark & \checkmark &  &  88.80$\pm$9.48 & 75.21$\pm$3.06 & 72.4$\pm$4.01  & 78.03$\pm$1.74 & 91.05$\pm$1.77 & 55.29$\pm$1.75 & 80.08$\pm$2.18 & 69.22$\pm$1.24  \\
~ & CDL & \checkmark& \checkmark& \checkmark & \textbf{89.94$\pm$8.76} & \textbf{76.10$\pm$2.80} & \textbf{74.90$\pm$3.70} & \textbf{79.08$\pm$1.86} & \textbf{92.05$\pm$2.14} & \textbf{56.51$\pm$2.32} & \textbf{80.96$\pm$1.29} & \textbf{70.83$\pm$1.13} \\
\midrule
\multirow{3}{*}{70\%} & CDL$_{\textbf{cl}}$ & & \checkmark &  & 89.39$\pm$7.83 & 75.03$\pm$4.34 & 73.00$\pm$4.78 & 81.44$\pm$1.97 & 90.55$\pm$1.48 & 55.27$\pm$1.42 & 81.04$\pm$1.54 & 69.34$\pm$1.16 \\
~& CDL$_{\textbf{ft}}$ &  \checkmark & \checkmark &  & 89.30$\pm$7.70 & 76.37$\pm$3.83 & 73.10$\pm$6.21 & 81.68$\pm$1.68 & 91.55$\pm$1.76 & 56.19$\pm$2.02 & 81.24$\pm$1.10 & 69.42$\pm$1.05  \\
~& CDL & \checkmark& \checkmark& \checkmark  & \textbf{89.91$\pm$7.30} & \textbf{77.27$\pm$3.62} & \textbf{74.90$\pm$5.63} & \textbf{82.36$\pm$1.52} & \textbf{92.35$\pm$1.63} & \textbf{56.65$\pm$1.87} & \textbf{82.36$\pm$1.72} & \textbf{71.06$\pm$1.17} \\
\bottomrule
\end{tabular}
\end{small}
\end{center}
\vskip -0.2in
\end{table*}

\subsubsection{Generalization Bound of the Distribution Divergence}
We analyze the generalization bound of the distribution divergence in ${\mathcal{L}_d}$ in Eq. \eqref{eq:lossdd}. The proof of Theorem \ref{th:bound} is provided in the supplementary material. According to Theorem \ref{th:bound}, ${L_d}$ has a specific lower bound in Eq. \eqref{eq:lossdd} if two conditions are satisfied, i.e., ${sim\left( {\mathbf{h}_i,\mathbf{h}_k^w} \right)}=0$ $\left( {1 \le k \le K} \right)$ and $K\ge {e^{ \frac{1}{\tau }}} - 1$. In particular, we have ${\mathcal{L}_d}  \ge 0$ if $K\ge {e^{ \frac{1}{\tau }}} - 1$. Therefore, a lower bound is theoretically guaranteed when minimizing ${\mathcal{L}}$ in Eq. \eqref{eq:ovloss}.

\begin{theorem} \label{th:bound}
Assume that there are $n_g$ graphs and $K$ negative samples for each graph, where ${sim\left( {\mathbf{h}_i,\mathbf{h}_k^w} \right)}=0$ $\left( {1 \le k \le K} \right)$. Given two conditional distributions $\mathbf{h}_i^w$ and $\mathbf{h}_i^s$ relative to $\mathbf{h}_i$, denoted as $ p\left( {\mathbf{h}_i^w|{\mathbf{h}_i}} \right)$ and $p\left( {\mathbf{h}_i^s|{\mathbf{h}_i}} \right)$, respectively, and the distribution divergence ${\mathcal{L}_d}$ in Eq. \eqref{eq:lossdd}, the following inequality holds:
\begin{equation*}
{\mathcal{L}_d} \ge \log \left( {K + 1} \right) - \frac{1}{\tau }
\end{equation*}
if $K$ satisfies the following condition, i.e.,
\begin{equation*}
K\ge {e^{ \frac{1}{\tau }}} - 1
\end{equation*}
where $\tau$ is a temperature parameter.
\end{theorem}

\begin{figure*}[htbp]
        \begin{minipage}[t]{0.33\linewidth}
        \centering
        \includegraphics[width=4.5cm]{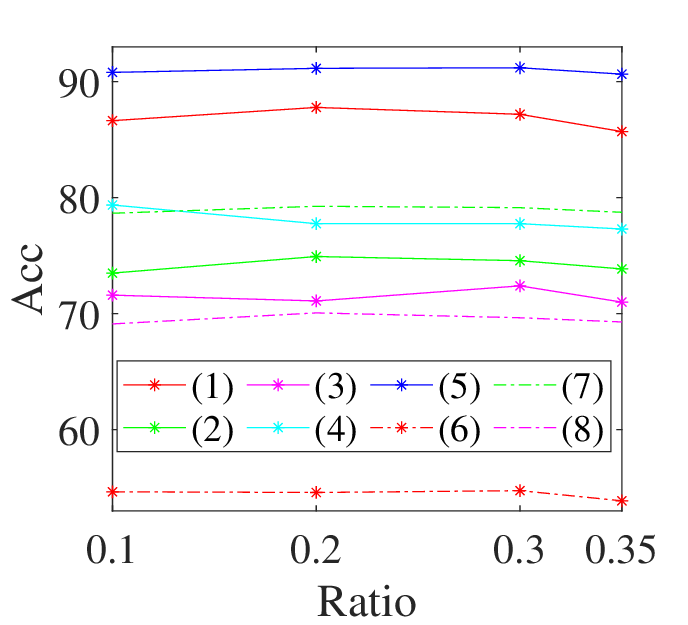}
        \caption*{(a)}
        \end{minipage}%
        \begin{minipage}[t]{0.33\linewidth}
        \centering
        \includegraphics[width=4.5cm]{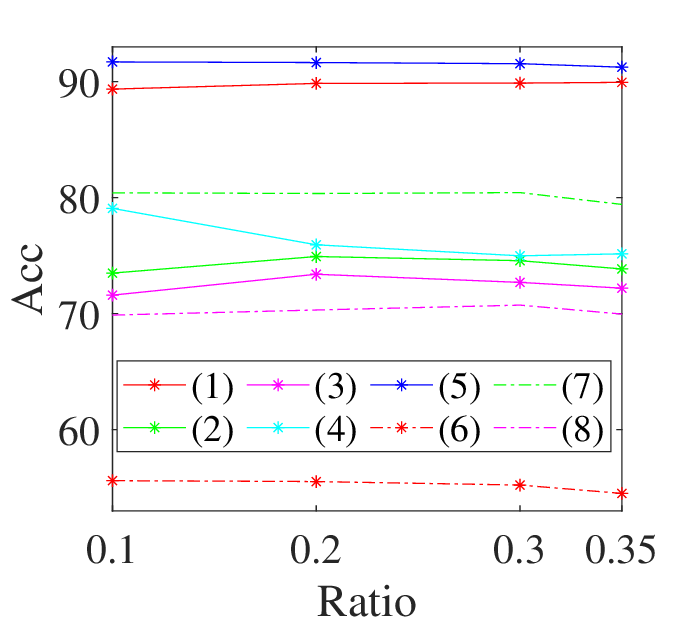}
        \caption*{(b)}
        \end{minipage}%
         \begin{minipage}[t]{0.33\linewidth}
        \centering
        \includegraphics[width=4.5cm]{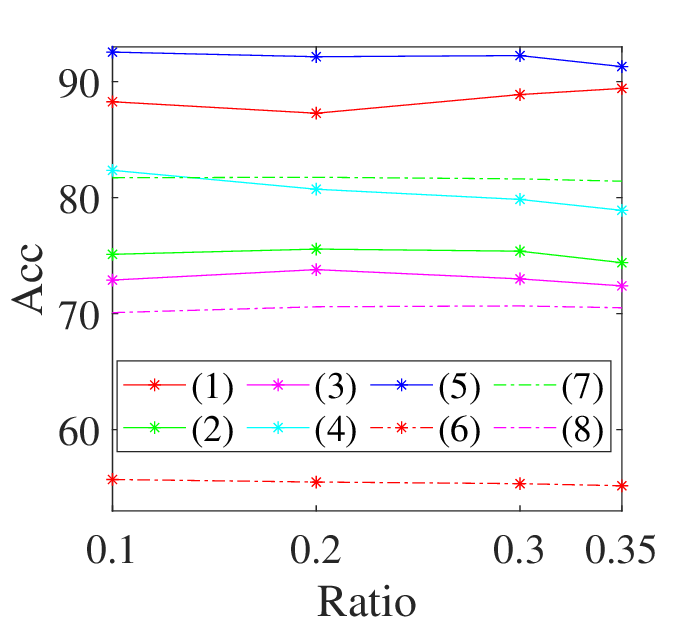}
        \caption*{(c)}
        \end{minipage}%
        \centering
        \caption{Graph classification results with different node masking ratios combinations across different percentages of training samples with the eight graph datasets. (a) 30\%, (b) 50\%, and  (c) 70\%. The datasets: (1) MUTAG, (2) PROTEINS, (3) IMDB-B, (4) NCI1, (5) RDT-B, (6) RDT-M5K, (7) COLLAB, and (8) GITHUB.}
         \label{fig:param:masking}
\end{figure*}

\section{Experiments}
\label{sec:exp}

\subsection{Experimental Settings}
We employed eight benchmark graph datasets from the TUDataset \cite{Morris2020TuData}, including MUTAG, PROTEINS, IMDB-B, NCI1, RDT-B, RDT-M5K, COLLA and GITHUB, for evaluation. The statistics of these datasets and the parameter settings are provided in the supplementary material. Following the experimental settings in \cite{Yue2022SSGC}, we evaluated all competing algorithms via 10-fold cross-validation. Each dataset was randomly shuffled and evenly divided into 10 parts. One part was employed as the test set, another as the validation set, and the remaining parts were used for training. The training set contains two types of graphs. Three different percentages of the dataset, 30\%, 50\% and 70\%, are selected from each graph set as labeled training graphs, while the labels of the remaining graphs are unknown during training. Classification accuracy was employed to evaluate the performance of all competing algorithms. We reported the average results and the standard deviations of the classification accuracy. A higher average accuracy on graph classification tasks indicates better preservation of the intrinsic semantic information contained in the graph-structured data. The source code of CDL is available at https://github.com/chenjie20/SSCDL.

To evaluate the performance of graph classification, we compared the results of the proposed CDL method with those of several state-of-the-art graph classification methods, including GCL \cite{You2020GCL}, GLIA \cite{Yue2022SSGC}, $\mathcal{G}$-Mixup \cite{Han2022GDA}, the graph contrastive masked autoencoder (GCMAE) \cite{Wang2024GCMAE} and graph reference distribution learning (GRDL) \cite{Wang2024GC}. We employed the GCN as a backbone in GCL. The graph classification results convey specific semantic information about the predicted labels of the graphs.

\subsubsection{Weak and Strong Augmentation Settings}
The same graph augmentation strategy, node attribute masking, was utilized for both the weak and strong augmentations applied to the graph datasets. Specifically, a relatively small masking ratio of the node attributes is regraded as slight perturbations and is treated as a weak augmentation of graph-structured data. The primary difference between the weak and strong augmentations lies in the node masking ratio. The masking ratio of the node attributes for the weak augmentation was selected from the set $\left\{ {0.1,0.2,0.3,0.35} \right\}$, whereas the ratio for the strong augmentation was set to be twice that of the weak augmentation. By adjusting the magnitude of node attribute corruption through the masking ratio, we aim to investigate different levels of semantic information distortion while preserving the integrity of the overall graph topology.

\subsection{Performance Evaluation}
The graph classification results for competing methods across the eight graph datasets are reported in Table \ref{tb:results}. The best and second-best results are highlighted in bold and underlined, respectively. Table \ref{tb:results} shows that the CDL method consistently outperforms the other competing methods. For example, the CDL method achieves performance improvements of approximately 2.11\%, 0.55\% and 0.49\% over the second-best method (GLIA) at labels ratios of 30\%, 50\% and 70\%, respectively, with the MUTAG dataset. The gap in the classification results between the CDL and GLIA is small when the label ratio is 50\% or 70\% with the MUTAG dataset, due to the small number of graphs in this dataset. However, the gap in the classification results widened with the other datasets as the number of graphs increased. Moreover, we observe the same advantages of the CDL method with different label ratios.

GCL and GLIA are two representative graph contrastive learning-based methods used in the experiments. The GLIA algorithm achieves better graph classification results than the other competing methods do, such as G-Mixup, GCMAE and GRDL. GCL also achieves satisfactory results in the experiments. This finding highlights the significant benefits of employing contrastive learning on graphs. The symbol '-' in Table \ref{tb:results} indicates an out-of-memory condition. GCMAE and GRDL often fail to perform graph classification for several of the large datasets, e.g., the RDT-M5K and GITHUB datasets, because of memory constraints.

\subsection{Ablation Study}
To investigate the impact of the loss components ${\mathcal{L}_s}$ in Eq. \eqref{eq:ovloss}, we performed ablation studies on graph classification tasks without the pretraining stage. In particular, we considered two scenarios in the experiments. We first employed the loss components ${\mathcal{L}_c}$ and ${\mathcal{L}_s}$ in Eq. \eqref{eq:ovloss}, which represents a semisupervised graph classification model with pretraining. This model retains only positive pairs of node representations for contrastive learning, which measures the similarity between the original features and the corresponding weakly augmented features. Then, we only conducted the fine-tuning stage of CDL, which involves the two loss components, ${\mathcal{L}_c}$ and ${\mathcal{L}_d}$, in Eq. \eqref{eq:ovloss}. The invariants of the methods corresponding to these two scenarios are referred to as CDL$_{\textbf{cl}}$ and CDL$_{\textbf{ft}}$, respectively.

Table \ref{tb:ablation} shows ablation study results for CDL$_{\textbf{cl}}$ and CDL$_{\textbf{ft}}$. On the one hand, CDL$_{\textbf{ft}}$ is superior to CDL$_{\textbf{cl}}$ in most cases, even without the pretraining stage. This finding further highlights the effectiveness of the conditional distribution in ${\mathcal{L}_s}$. On the other hand, CDL achieves superior graph classification results compared with those of CDL$_{\textbf{ft}}$. This finding indicates that pretraining helps preserve intrinsic semantic information, which significantly improves the performance of the proposed CDL method. These experimental results demonstrate the effectiveness of our proposed semisupervised training scheme for graph classification.

\subsection{Empirical Investigation}
We empirically investigated the impact of varying node masking ratios on the performance of the proposed CDL method. The node masking ratio for weak augmentation was selected from the set $\left\{ {0.1,0.2,0.3,0.35} \right\}$. The node masking ratio for strongly augmented features is twice that of weak augmentation. Fig. \ref{fig:param:masking} shows the graph classification results with different node masking ratios across different percentages of training samples with eight benchmark graph datasets. The graph classification result often increases when the node masking ratio gradually increases from 0.1 to 0.3 with most datasets. This finding demonstrates that conditional distribution learning can effectively support graph representation learning. Additionally, the optimal range for the node masking ratio is relatively narrow, making it easier to determine this parameter in practice. In contrast, the graph classification result often decreases when the node masking ratio increases to 0.35. The primary reason for the decline in graph classification performance is the severe degradation of graph structure information in the strongly augmented view, where the node masking ratio is 0.7.

\section{Conclusion}
\label{sec:conclusion}
In this paper, we introduce the CDL method for semisupervised graph classification tasks. The CDL model is designed to align the conditional distributions of weakly and strongly augmented features over the original features. The proposed CDL method leverages the diversity and abundance provided by graph-structured data augmentations while preserving intrinsic semantic information. Moreover, CDL resolves the conflict between the message-passing mechanism of GNNs and the contrastive learning of negative pairs within intraviews. Additionally, we present a semisupervised learning scheme consisting of pretraining and fine-tuning stages. Extensive experimental results with various graph datasets demonstrate the superior performance of the CDL method.



\end{document}